# Playing for Data: Ground Truth from Computer Games

Stephan R. Richter[*1]    Vibhav Vineet[*2]    Stefan Roth[1]    Vladlen Koltun[2]

[1]TU Darmstadt        [2]Intel Labs

**Abstract.** Recent progress in computer vision has been driven by high-capacity models trained on large datasets. Unfortunately, creating large datasets with pixel-level labels has been extremely costly due to the amount of human effort required. In this paper, we present an approach to rapidly creating pixel-accurate semantic label maps for images extracted from modern computer games. Although the source code and the internal operation of commercial games are inaccessible, we show that associations between image patches can be reconstructed from the communication between the game and the graphics hardware. This enables rapid propagation of semantic labels within and across images synthesized by the game, with no access to the source code or the content. We validate the presented approach by producing dense pixel-level semantic annotations for 25 thousand images synthesized by a photorealistic open-world computer game. Experiments on semantic segmentation datasets show that using the acquired data to supplement real-world images significantly increases accuracy and that the acquired data enables reducing the amount of hand-labeled real-world data: models trained with game data and just $\frac{1}{3}$ of the CamVid training set outperform models trained on the complete CamVid training set.

## 1 Introduction

Recent progress in computer vision has been driven by high-capacity models trained on large datasets. Image classification datasets with millions of labeled images support training deep and highly expressive models [24]. Following their success in image classification, these models have recently been adapted for detailed scene understanding tasks such as semantic segmentation [28]. Such semantic segmentation models are initially trained for image classification, for which large datasets are available, and then fine-tuned on semantic segmentation datasets, which have fewer images.

We are therefore interested in creating very large datasets with pixel-accurate semantic labels. Such datasets may enable the design of more diverse model architectures that are not constrained by mandatory pre-training on image classification. They may also substantially increase the accuracy of semantic segmentation models, which at present appear to be limited by data rather than capacity. (For example, the top-performing semantic segmentation models on the PASCAL VOC leaderboard all use additional external sources of pixelwise labeled data for training.)

Creating large datasets with pixelwise semantic labels is known to be very challenging due to the amount of human effort required to trace accurate object boundaries.

---

\* Authors contributed equally

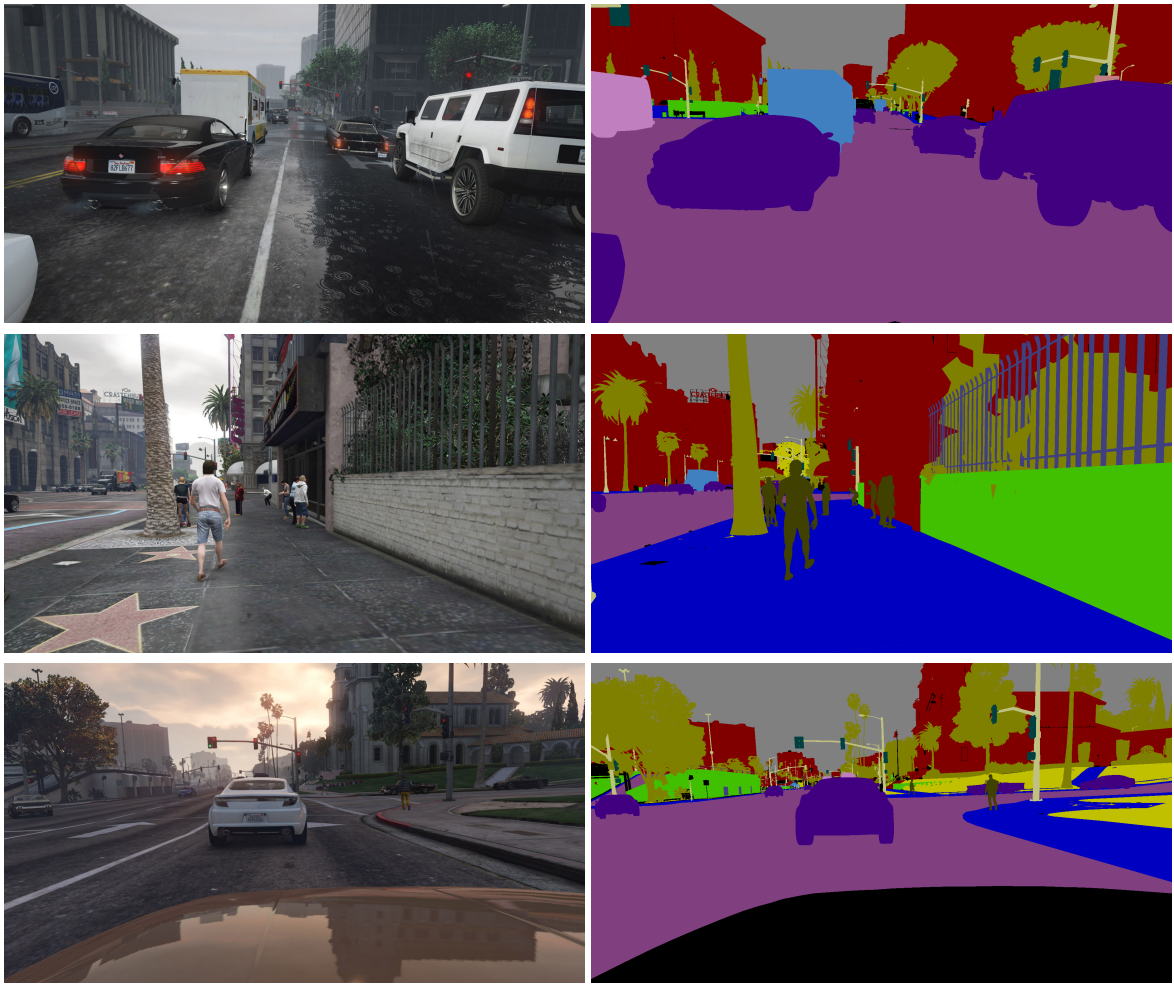

**Fig. 1.** Images and ground-truth semantic label maps produced by the presented approach. Left: images extracted from the game Grand Theft Auto V. Right: semantic label maps. The color coding is defined in Fig. 4.

High-quality semantic labeling was reported to require 60 minutes per image for the CamVid dataset [8] and 90 minutes per image for the Cityscapes dataset [11]. Due to the substantial manual effort involved in producing pixel-accurate annotations, semantic segmentation datasets with precise and comprehensive label maps are orders of magnitude smaller than image classification datasets. This has been referred to as the "curse of dataset annotation" [50]: the more detailed the semantic labeling, the smaller the datasets.

In this work, we explore the use of commercial video games for creating large-scale pixel-accurate ground truth data for training semantic segmentation systems. Modern open-world games such as Grand Theft Auto, Watch Dogs, and Hitman feature extensive and highly realistic worlds. Their realism is not only in the high fidelity of material appearance and light transport simulation. It is also in the content of the game worlds: the layout of objects and environments, the realistic textures, the motion of vehicles and autonomous characters, the presence of small objects that add detail, and the interaction between the player and the environment.

The scale, appearance, and behavior of these game worlds are significant advantages over open-source sandboxes that lack this extensive content. However, detailed semantic annotation of images from off-the-shelf games is a challenge because the internal operation and content of the game are largely inaccessible. We show that this can be overcome by a technique known as detouring [19]. We inject a wrapper between the game and the operating system, allowing us to record, modify, and reproduce rendering commands. By hashing distinct rendering resources – such as geometry, textures, and shaders – communicated by the game to the graphics hardware, we generate object signatures that persist across scenes and across gameplay sessions. This allows us to create pixel-accurate object labels without tracing boundaries. Crucially, it also enables propagating object labels across time and across instances that share distinctive resources.

Using the presented approach, we have created pixel-level semantic segmentation ground truth for 25 thousand images extracted from the game Grand Theft Auto V. The labeling process was completed in only 49 hours. Our labeling speed was thus roughly three orders of magnitude faster than for other semantic segmentation datasets with similar annotation density. The pixel-accurate propagation of label assignments through time and across instances led to a rapid acceleration of the labeling process: average annotation time per image decreased sharply during the process because new object labels were propagated across images. This is in contrast to prior labeling interfaces, in which annotation speed does not change significantly during the labeling process, and total labeling costs scale linearly with dataset size. Annotating the presented dataset with the approach used for CamVid or Cityscapes [8, 11] would have taken at least 12 person-years. Three of the images we have labeled are shown in Fig. 1.

To evaluate the utility of using game data for training semantic segmentation systems, we used label definitions compatible with other datasets for urban scene understanding [8, 11, 13, 50]. We conducted extensive experiments to evaluate the effectiveness of models trained with the acquired data. The experimental results show that using the acquired data to supplement the CamVid and KITTI training sets significantly increases accuracy on the respective datasets. In addition, the experiments demonstrate that the acquired data can reduce the need for expensive labeling of real-world images: models trained with game data and just $\frac{1}{3}$ of the CamVid training set outperform models trained on the complete CamVid training set.

## 2 Related Work

Synthetic data has been used for decades to benchmark the performance of computer vision algorithms. The use of synthetic data has been particularly significant in evaluating optical flow estimation due to the difficulty of obtaining accurate ground-truth flow measurements for real-world scenes [6, 7, 18, 32]. Most recently, the MPI-Sintel dataset has become a standard benchmark for optical flow algorithms [9] and has additionally yielded ground-truth data for depth estimation and bottom-up segmentation. Synthetic scenes have been used for evaluating the robustness of image features [21] and for benchmarking the accuracy of visual odometry [16]. Renderings of object models have been used to analyze the sensitivity of convolutional network features [5]. In contrast to this line of work, we use synthetic data not for benchmarking but for training a vision system, and tackle the challenging problem of semantic segmentation.

Rendered depth maps of parametric models have been used prominently in training leading systems for human pose estimation and hand tracking [41, 42]. 3D object models are also increasingly used for training representations for object detection and object pose estimation [4, 26, 30, 34, 35, 44]. Renderings of 3D object models have been used to train shape-from-shading algorithms [37] and convolutional networks for optical flow estimation [12]. Renderings of entire synthetic environments have been proposed for training convolutional networks for stereo disparity and scene flow estimation [31]. Our work is different in two ways. First, we tackle the problem of semantic segmentation, which involves both recognition and perceptual grouping [17, 23, 28, 43]. Second, we obtain data not by rendering 3D object models or stylized environments, but by extracting photorealistic imagery from a modern open-world computer game with high-fidelity content.

Computer games – and associated tools such as game engines and level editors – have been used a number of times in computer vision research. Development tools accompanying the game Half Life 2 were used for evaluating visual surveillance systems [46]. These tools were subsequently used for creating an environment for training high-performing pedestrian detectors [29, 49, 51]. And an open-source driving simulator was used to learn mid-level cues for autonomous driving [10]. In contrast to these works, we deal with the problem of semantic image segmentation and demonstrate that data extracted from an unmodified off-the-shelf computer game with no access to the source code or the content can be used to substantially improve the performance of semantic segmentation systems.

Somewhat orthogonal to our work is the use of indoor scene models to train deep networks for semantic understanding of indoor environments from depth images [15, 33]. These approaches compose synthetic indoor scenes from object models and synthesize depth maps with associated semantic labels. The training data synthesized in these works provides depth information but no appearance cues. The trained models are thus limited to analyzing depth maps. In contrast, we show that modern computer games can be used to increase the accuracy of state-of-the-art semantic segmentation models on challenging real-world benchmarks given regular color images only.

## 3   Breaking the Curse of Dataset Annotation

Extracting images and metadata from a game is easy if the source code and content are available [10, 14]. Open-source games, however, lack the extensive, detailed, and realistic content of commercial productions. In this section, we show that rapid semantic labeling can be performed on images generated by off-the-shelf games, without access to their source code or content. We then demonstrate the presented approach by producing pixel-accurate semantic annotations for 25 thousand images from the game Grand Theft Auto V. (The publisher of Grand Theft Auto V allows non-commercial use of footage from the game as long as certain conditions are met, such as not distributing spoilers [38].)

## 3.1  Data acquisition

**A brief introduction to real-time rendering.** To present our approach to data acquisition, we first need to review some relevant aspects of the rendering pipeline used in modern computer games [2]. Modern real-time rendering systems are commonly based on deferred shading. Geometric resources are communicated to the GPU to create a depth buffer and a normal buffer. Maps that specify the diffuse and specular components of surface reflectance are communicated to create the diffuse and specular buffers. Buffers that collect such intermediate products of the rendering pipeline are called G-buffers. Illumination is applied to these G-buffers, rather than to the original scene components [40]. This decoupled processing of geometry, reflectance properties, and illumination significantly accelerates the rendering process. First, shading does not need to be performed on elements that are subsequently culled by the geometry pipeline. Second, shading can be performed much more efficiently on G-buffers than on an unstructured stream of objects. For these reasons, deferred shading has been widely adopted in high-performance game engines.

**Extracting information from the rendering pipeline.** How can this pipeline be employed in creating ground-truth semantic labels if we don't have access to the game's code or content? The key lies in the game's communication with the graphics hardware. This communication is structured. Specifically, the game communicates resources of different types, including geometric meshes, texture maps, and shaders. The game then specifies how these resources should be combined to compose the scene. The content of these resources persists through time and across gameplay sessions. By tracking the application of resources to different scene elements, we can establish associations between these scene elements.

Our basic approach is to intercept the communication between the game and the graphics hardware. Games communicate with the hardware through APIs such as OpenGL, Direct3D, or Vulkan, which are provided via dynamically loaded libraries. To initiate the use of the hardware, a game loads the library into its application memory. By posing as the graphics library during this loading process, a wrapper to the library can be injected and all subsequent communication between the game and the graphics API can be monitored and modified. This injection method is known as detouring [19] and is commonly used for patching software binaries and achieving program interoperability. It is also used by screen-capturing programs and off-the-shelf graphics debugging tools such as RenderDoc [22] and Intel Graphics Performance Analyzers [20]. To perform detouring, a wrapper needs to implement all relevant interfaces and forward calls to the original library. We implemented a wrapper for the DirectX 9 API and used RenderDoc for wrapping Direct3D 11. We successfully tested these two implementations on three different rendering engines used in AAA computer games. By intercepting all communication with the graphics hardware, we are able to monitor the creation, modification, and deletion of resources used to specify the scene and synthesize an image.

We now focus on the application of our approach to the game Grand Theft Auto V (GTA5), although much of the discussion applies more broadly. To collect data, we used RenderDoc to record every 40th frame during GTA5 gameplay. Being a debugger for applications, RenderDoc can be configured to record all calls of an application to the graphics API with their respective arguments and allows detailed inspection of the arguments. Since RenderDoc is scriptable and its source code is available as well, we modified it to automatically transform recorded data into a format that is suitable for annotation.

Specifically, the wrapper saves all information needed to reproduce a frame. The frames are then processed in batch after a gameplay session to extract all information needed for annotation. (This separate processing requires about 30 seconds per frame.) Annotation of large batches of collected and processed frames is performed later in an interactive interface that uses the extracted information to support highly efficient annotation (Fig. 3). In the following paragraphs, we discuss several challenges that had to be addressed:

1. Identify function calls that are relevant for rendering objects into the set of G-buffers that we are interested in.
2. Create persistent identities for resources that link their use across frames and across gameplay sessions.
3. Organize and store resource identities to support rapid annotation in a separate interactive interface.

**Identifying relevant function calls.** To identify rendering passes, RenderDoc groups function calls into common rendering passes based on predefined heuristics. We found that strictly grouping the calls by the G-buffers that are assigned as render targets works more reliably. That way, we identify the main pass that processes the scene geometry and updates the albedo, surface normal, stencil, and depth buffers as well as the rendering passes that draw the head-up display on top of the scene image. GTA5 applies post-processing effects such as camera distortion to the rendered image before displaying it. To preserve the association of object information extracted from the main pass with pixels in the final image and to bypass drawing the head-up display (HUD), we omit the camera distortion and subsequent HUD passes.

**Identifying resources.** To propagate labels across images, we need to reliably identify resources used to specify different scene elements. When the same mesh is used in two different frames to specify the shape of a scene element, we want to reliably recognize that it is the same mesh. During a single gameplay session, a resource can be recognized by its location in memory or by the ID used by the application to address this resource. However, the next time the game is launched, memory locations and IDs associated with resources will be different. To recognize resources across different gameplay sessions, we instead hash the associated memory content. We use a non-cryptographic hash function [3] to create a 128-bit key for the content of the memory occupied by the resource. This key is stored and is used to identify the resource in different frames. Thus, for each recorded frame, we create a lookup table to map the volatile resource IDs to persistent hash keys.

**Formatting for annotation.** Although we can now identify and associate resources that are being used to create different frames, we have not yet associated these resources with pixels in the rendered images. We want to associate each mesh, texture, and shader with their footprint in each rendered image. One way to do this would be to step through the rendering calls and perform pixel-level comparisons of the content of each G-buffer before and after each call. However, this is unreliable and computationally expensive. Instead, we perform two complete rendering passes instead of one and produce two distinct images. The first rendering pass produces the color image and the associated buffers as described above: this is the conventional rendering pass performed by the game. The second rendering pass is used to encode IDs into pixels, such that after this pass each pixel stores the resource IDs for the mesh, texture, and shader that specify the scene element imaged at that pixel. For this second rendering pass, we replace all the shaders with our own custom shader that encodes the resource IDs of textures, meshes, and original shaders into colors and writes them into four render targets. Four render targets with three 8-bit color channels each provide us with 96 bits per pixel, which we use to store three 32-bit resource IDs: one for the mesh, one for the texture, one for the shader. In a subsequent processing stage, we read off these 32-bit IDs, which do not persist across frames, and map them to the persistent 128-bit hash keys created earlier.

### 3.2   Semantic labeling

For each image extracted from the game, the pipeline described in Section 3.1 produces a corresponding resource ID map. For each pixel, this ID map identifies the mesh, texture, and shader that were used by the surface imaged at that pixel. These IDs are persistent: the same resource is identified by the same ID in different frames. This is the data used by the annotation process.

**Patch decomposition.** We begin by automatically decomposing each image into patches of pixels that share a common ⟨mesh, texture, shader⟩ combination (henceforth, MTS). Fig. 2 shows an image from the game and the resulting patches. The patches are fine-grained and each object is typically decomposed into multiple patches. Furthermore, a given patch is likely to be contained within a single object. A given mesh may contain multiple objects (a building and an adjacent sidewalk), a texture may be used on objects from different semantic classes (car and truck), and a shader may likewise be applied on semantically distinct objects, but an MTS combination is almost always used within a single object type.

The identified patches are thus akin to superpixels [36], but have significant advantages over superpixels generated by a bottom-up segmentation algorithm. First, they are associated with the underlying surfaces in the scene, and patches that depict the same surface in different images are linked. Second, boundaries of semantic classes in an image coincide with patch boundaries. There is no need for a human annotator to delineate object boundaries: pixel-accurate label maps can be created simply by grouping patches. Third, as we shall see next, the metadata associated with each patch can be used to propagate labels even across object instances that do not share the same MTS.

**Association rule mining.** So far we have required that two patches share a complete MTS combination to be linked. However, requiring that the mesh, texture, and shader all be identical is sometimes too conservative: there are many cases in which just one or two resources are sufficient to uniquely identify the semantic class of a patch. For example, a car mesh is highly unlikely to be used for anything but a car. Instead of specifying such cases by hand, we discover them automatically during the labeling process via association rule mining [1].

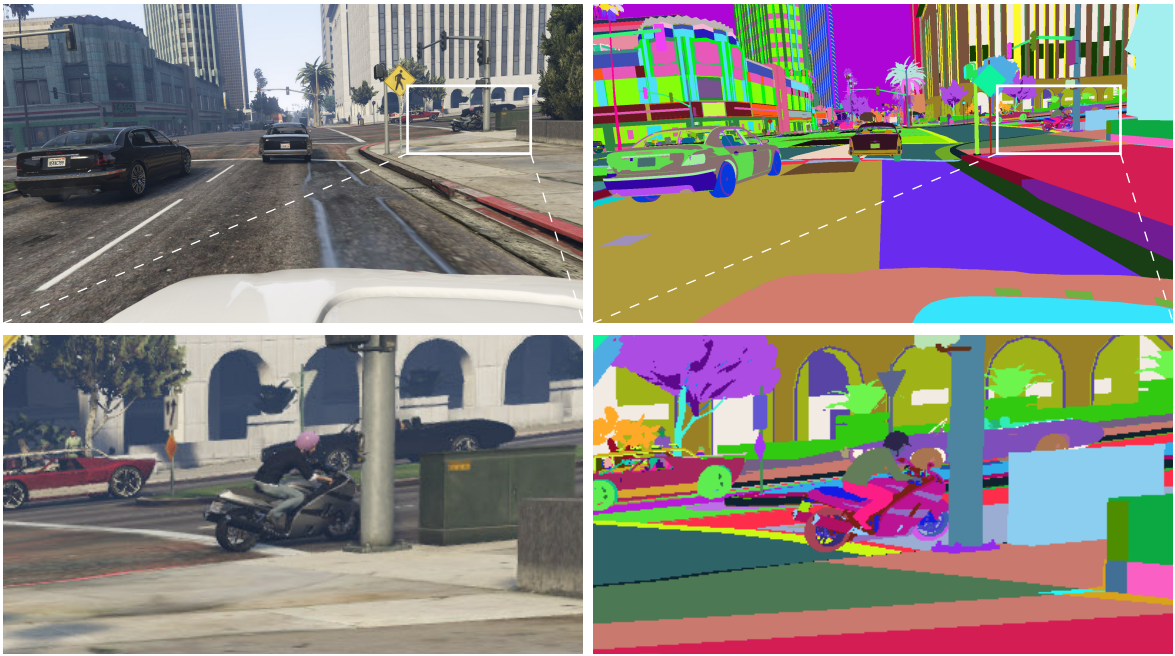

**Fig. 2.** Illustration of the patches used as atomic units during the annotation process. Top left: one of the images in our dataset. Top right: patches constructed by grouping pixels that share a common MTS combination. Different patches are filled with different colors for illustration. Bottom row: partial magnifications of the top images.

During the annotation process, statistical regularities in associations between resources and semantic labels are detected. When sufficient evidence is available for a clear association between a resource and a semantic class, a rule is automatically created that labels other patches that use this resource by the associated class. This further speeds up annotation by propagating labels not just to observations of the same surface in the scene at different times, but also across different objects that use a distinctive resource that clearly identifies their semantic class.

**Annotation process.** We use a simple interactive interface that supports labeling patches by clicking. The interface is shown in Fig. 3. Labeled areas are tinted by the color of their semantic class. (The color definitions are shown in Fig. 4.) The annotator selects a semantic class and then clicks on a patch that has not yet been labeled. In Fig. 3(left), four patches are unlabeled: part of a sidewalk, a fire hydrant, and two patches on the building. They are labeled in 14 seconds to yield the complete label map shown in Fig. 3(right).

Labeling the very first image still takes time due to the granularity of the patches. However, the annotation tool automatically propagates labels to all patches that share the same MTS in all images, as well as other patches that are linked by distinctive resources identified by association rule mining. As the annotation progresses, more and more patches in all images are pre-labeled. The annotation time per image decreases during the process: the more images have been annotated, the faster the labeling of each new image is. Our annotation tool only presents an image for annotation if more than 3% of the image area has not been pre-labeled automatically by propagating labels from other frames. In this way, only a fraction of the images has to be explicitly handled by the annotator.

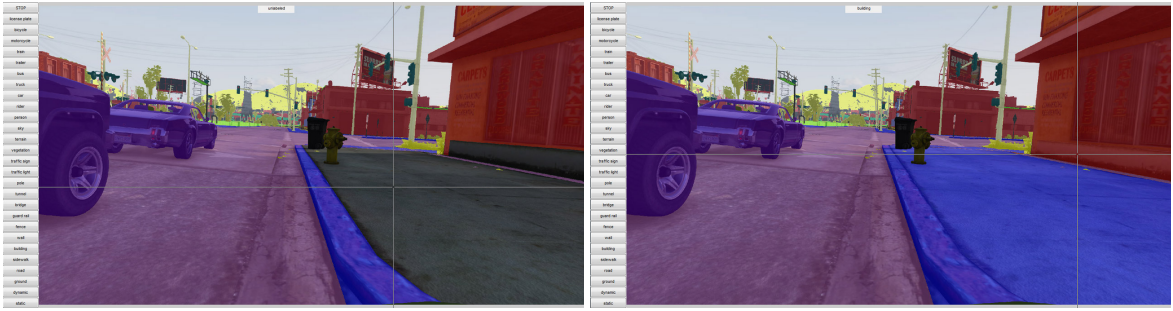

**Fig. 3.** Annotation interface. Labeled patches are tinted by the color of their semantic class. The annotator selects a semantic class and applies it to a patch with a single click. Left: an intermediate state with four patches yet unlabeled. Right: a complete labeling produced 14 seconds later.

Labeling each pixel in every image directly would be difficult even with our single-click interface because distant or heavily occluded objects are often hard to recognize. The label propagation ensures that even if a patch is left unlabeled in a given image because it is small or far away, it will likely be labeled eventually when the underlying surface is seen more closely and its assigned label is propagated back.

### 3.3 Dataset and analysis

We extracted 24,966 frames from GTA5. Each frame has a resolution of $1914 \times 1052$ pixels. The frames were then semantically labeled using the interface described in Section 3.2. The labeling process was completed in 49 hours. In this time, 98.3% of the pixel area of the extracted images was labeled with corresponding semantic classes. Classes were defined to be compatible with other semantic segmentation datasets for outdoor scenes [8, 11, 39, 50]. The distribution of classes in our dataset is shown in Fig. 4.

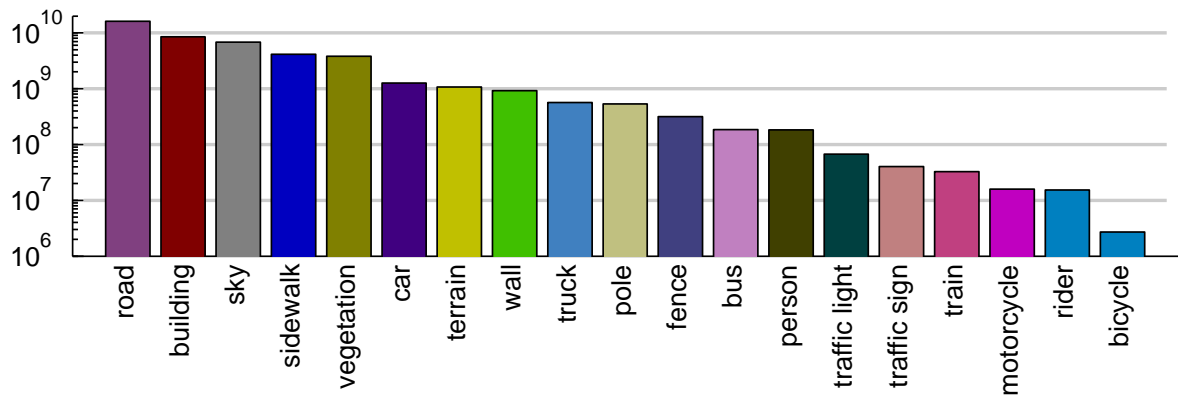

**Fig. 4.** Number of annotated pixels per class in our dataset. Note the logarithmic scale.

General statistics for the dataset are summarized in Table 1. Our dataset is roughly two orders of magnitude larger than CamVid [8] and three orders of magnitude larger than semantic annotations created for the KITTI dataset [13, 39]. The average annotation time for our dataset was 7 seconds per image: 514 times faster than the per-image annotation time reported for CamVid [8] and 771 times faster than the per-image annotation time for Cityscapes [11].

**Table 1.** Comparison of densely labeled semantic segmentation datasets for outdoor scenes. We achieve a three order of magnitude speed-up in annotation time, enabling us to densely label tens of thousands of high-resolution images.

|  | #pixels [$10^9$] | annotation density [%] | annotation time [sec/image] | annotation speed [pixels/sec] |
|---|---|---|---|---|
| GTA5 | 50.15 | 98.3 | 7 | 279,540 |
| Cityscapes (fine) [11] | 9.43 | 97.1 | 5400 | 349 |
| Cityscapes (coarse) [11] | 26.0 | 67.5 | 420 | 3095 |
| CamVid [8] | 0.62 | 96.2 | 3,600 | 246 |
| KITTI [39] | 0.07 | 98.4 | N/A | N/A |

**Label propagation.** The label propagation mechanisms significantly accelerated annotation time. Specific MTS combinations labeled during the process cover 73% of the cumulative pixel area in the dataset. Only a fraction of that area was directly labeled by the annotator; most labels were propagated from other images. Patches covered by learned association rules account for 43% of the dataset. During the annotation process, 2,295 association rules were automatically constructed. The union of the areas covered by labeled MTS combinations and learned association rules accounts for the 98.3% annotation density of our dataset.

Define the pre-annotated area of an image to be the set of patches that are pre-labeled before the annotator reaches that image. The patches in the pre-annotated area are pre-labeled by label propagation across patches that share the same MTS and via learned association rules. For each image, we can measure the size of the pre-annotated area relative to the whole frame. This size is $0\%$ if none of the image area is pre-annotated (e.g., for the very first image processed by the annotator) and $100\%$ if the entirety of the image area is already annotated by the time the annotator reaches this image. In a conventional annotation process used for datasets such as CamVid and Cityscapes, the pre-annotated area is a constant $0\%$ for all images. The pre-annotated area for images handled during our labeling process is plotted in Fig. 5. $98.7\%$ of the frames are more than $90\%$ pre-annotated by the time they are reached by the human annotator.

**Diversity of the collected data.** We also analyze the diversity of the images extracted from the game world. The effectiveness of label propagation may suggest that the collected images are visually uniform. This is not the case. Fig. 6 shows the distribution of the number of frames in which MTS combinations in the dataset occur. As shown in the figure, 26.5% of the MTS combinations only occur in a single image in the collected dataset. That is, more than a quarter of the MTS combinations observed in the 25 thousand collected images are only observed in a single image each. The median number of frames in which an MTS occurs is 4: that is, most of the MTS combinations are only seen in 4 images or less out of the 25 thousand collected images. This indicates the high degree of variability in the collected dataset.

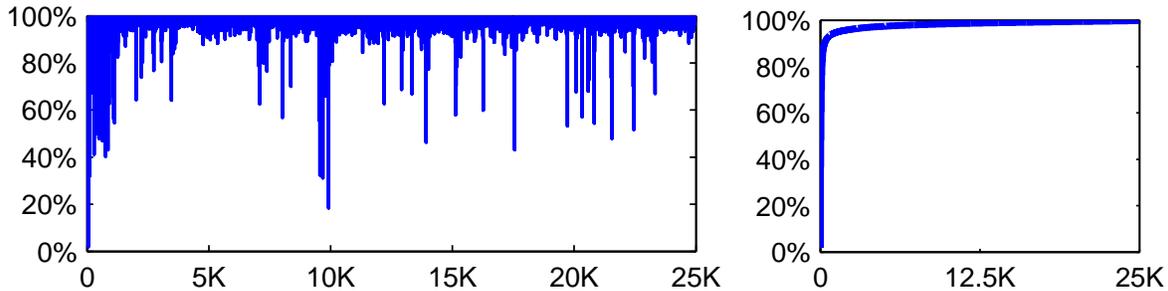

**Fig. 5.** Effect of label propagation. For each frame, the plots show the fraction of image area that is pre-annotated by the time the human annotator reaches that frame. On the left, the frames are arranged in the order they are processed; on the right, the frames are sorted by magnitude of pre-annotated area. Only 333 frames (1.3% of the total) are less than 90% pre-annotated by the time they are reached by the human annotator.

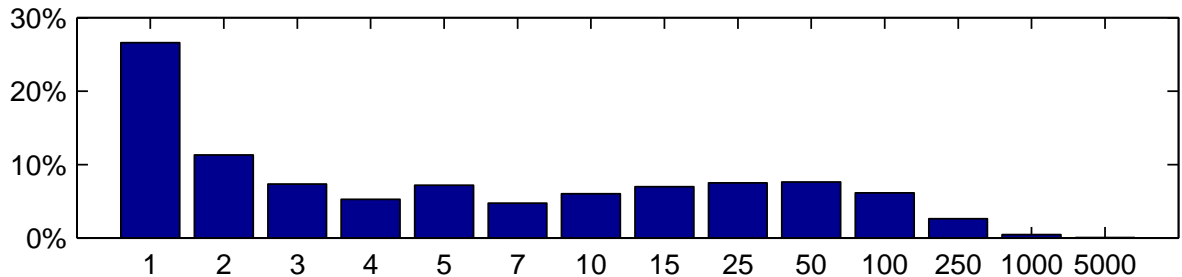

**Fig. 6.** Distribution of the number of frames in which an MTS combination occurs. 26.5% of the MTS combinations only occur in a single image each and most of the MTS combinations are only seen in 4 images or less.

In addition, Fig. 7 shows 20 randomly sampled images from our dataset. As the figure demonstrates, the images in the collected dataset are highly variable in their content and layout.

## 4    Semantic Segmentation

We now evaluate the effectiveness of using the acquired data for training semantic segmentation models. We evaluate on two datasets for semantic segmentation of outdoor scenes: CamVid [8] and KITTI [13, 39]. As our semantic segmentation model, we use the front-end prediction module of Yu and Koltun [52]. Our training procedure consists of two stages. In the first stage, we jointly train on real and synthetic data using mini-batch stochastic gradient descent with mini-batches of 8 images: 4 real and 4 synthetic. 50K iterations are performed with a learning rate of $10^{-4}$ and momentum 0.99. The crop size is $628 \times 628$ and the receptive field is $373 \times 373$ pixels. In the second stage, we fine-tune for 4K iterations on real data only, using the same parameters as in the first stage.

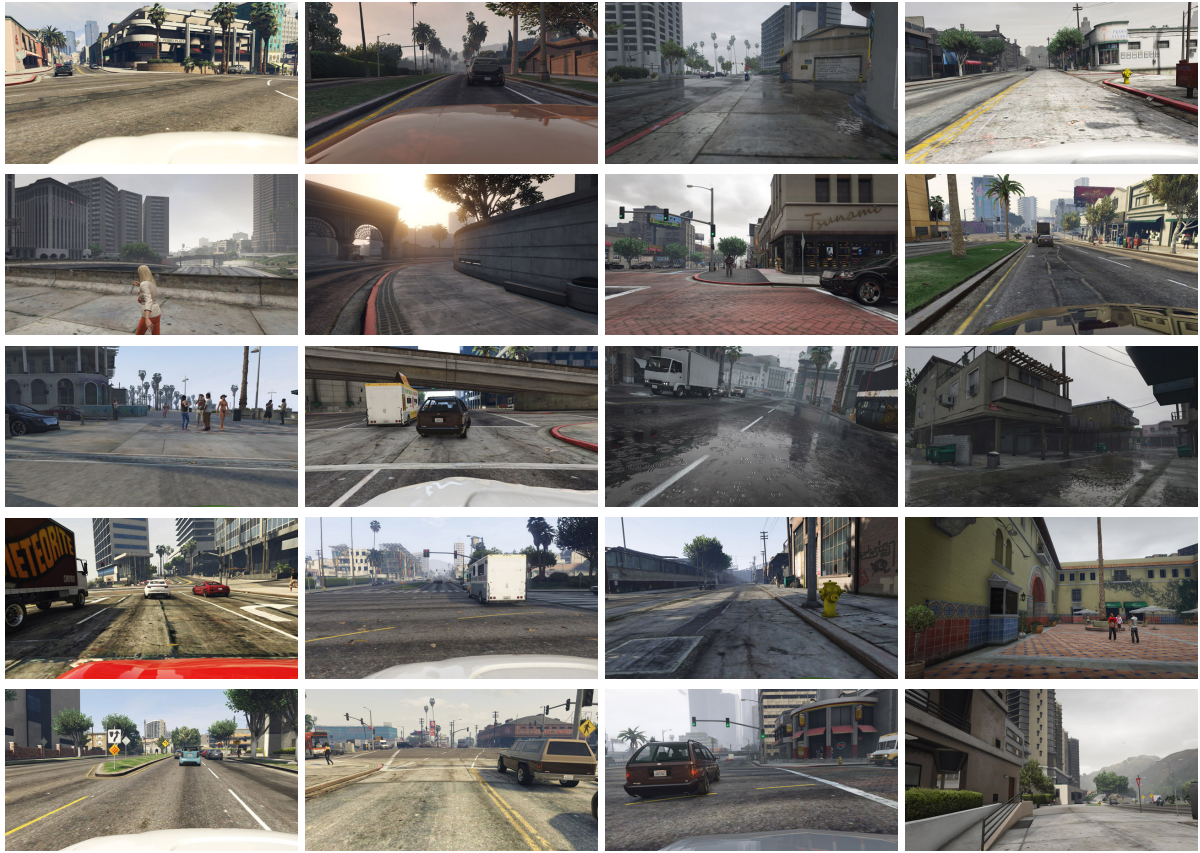

**Fig. 7.** Randomly sampled images from the collected dataset, illustrating its diversity.

### 4.1   CamVid dataset

We begin with experiments on the CamVid dataset. For ease of comparison to prior work, we adopted the training and test setup of Sturgess et al. [45], which has become standard for the CamVid dataset. This setup has 11 semantic classes and a split of the dataset into 367 training, 100 validation, and 233 test images.

The main results are summarized in Table 2. The table shows that using the synthetic data during training increases the mean IoU by 3.9 percentage points. In addition, we used the full set of synthetic images and varied the proportion of real images in the training set. The results show that when we train on $\frac{1}{3}$ of the CamVid training set along with the game data, we surpass the accuracy achieved when training on the full CamVid training set without game data. This suggests that the presented approach to acquiring and labeling synthetic data can significantly reduce the amount of hand-labeled real-world images required for training semantic segmentation models.

**Table 2.** Controlled experiments on the CamVid dataset. Training with the full CamVid training set augmented by the synthetic images increases the mean IoU by 3.9 percentage points. Synthetic images also allow reducing the amount of labeled real-world training data by a factor of 3.

| real images | 100% | - | 25% | 33% | 50% | 100% |
|---|---|---|---|---|---|---|
| synthetic images (all) | - | 100% | ✓ | ✓ | ✓ | ✓ |
| mean IoU | 65.0 | 43.6 | 63.9 | 65.2 | 66.5 | **68.9** |

Table 3 provides a comparison of our results to prior work on semantic segmentation on the CamVid dataset. Our strongest baseline is the state-of-the-art system of Kundu et al. [25], which used a bigger ConvNet and analyzed whole video sequences. By using synthetic data during training, we outperform this baseline by 2.8 percentage points, while considering individual frames only.

**Table 3.** Comparison to prior work on the CamVid dataset. We outperform the state-of-the-art system of Kundu et al. by 2.8 percentage points without utilizing temporal cues.

| Method | Building | Tree | Sky | Car | Sign | Road | Pedestrian | Fence | Pole | Sidewalk | Bicyclist | mean IoU |
|---|---|---|---|---|---|---|---|---|---|---|---|---|
| SuperParsing [47] | 70.4 | 54.8 | 83.5 | 43.3 | 25.4 | 83.4 | 11.6 | 18.3 | 5.2 | 57.4 | 8.9 | 42.0 |
| Liu and He [27] | 66.8 | 66.6 | 90.1 | 62.9 | 21.4 | 85.8 | 28 | 17.8 | 8.3 | 63.5 | 8.5 | 47.2 |
| Tripathi et al. [48] | 74.2 | 67.9 | 91 | 66.5 | 23.6 | 90.7 | 26.2 | 28.5 | 16.3 | 71.9 | 28.2 | 53.2 |
| Yu and Koltun [52] | 82.6 | 76.2 | 89.9 | 84.0 | 46.9 | 92.2 | 56.3 | 35.8 | 23.4 | 75.3 | 55.5 | 65.3 |
| Kundu et al. [25] | **84** | **77.2** | **91.3** | **85.6** | 49.9 | 92.5 | **59.1** | 37.6 | 16.9 | 76.0 | 57.2 | 66.1 |
| Our result | **84.4** | **77.5** | **91.1** | **84.9** | **51.3** | **94.5** | 59 | **44.9** | **29.5** | **82** | **58.4** | **68.9** |

### 4.2 KITTI dataset

We have also performed an evaluation on the KITTI semantic segmentation dataset. The results are reported in Table 4. We use the split of Ros et al. [39], which consists of 100 training images and 46 test images. We compare against several baselines for which the authors have either provided results on this dataset or released their code. The model trained with game data outperforms the model trained without game data by 2.6 percentage points.

## 5 Discussion

We presented an approach to rapidly producing pixel-accurate semantic label maps for images synthesized by modern computer games. We have demonstrated the approach by creating dense pixel-level semantic annotations for 25 thousand images extracted from a realistic open-world game. Our experiments have shown that data created with the presented approach can increase the performance of semantic segmentation models on real-world images and can reduce the need for expensive conventional labeling.

**Table 4.** Results on the KITTI dataset. Training with game data yields a 2.6 percentage point improvement over training without game data.

| Method | Building | Tree | Sky | Car | Sign | Road | Pedestrian | Fence | Pole | Sidewalk | Bicyclist | mean IoU |
|---|---|---|---|---|---|---|---|---|---|---|---|---|
| Ros et al. [39] | 71.8 | 69.5 | **84.4** | 51.2 | 4.2 | 72.4 | 1.7 | 32.4 | 2.6 | 45.3 | 3.2 | 39.9 |
| Tripathi et al. [48] | 75.1 | 74.0 | **84.4** | 61.8 | 0 | 75.4 | 0 | 1.0 | 2.2 | 37.9 | 0 | 37.4 |
| Yu and Koltun [52] | 84.6 | **81.1** | 83 | 61.4 | 41.8 | **92.9** | 4.6 | **47.1** | 35.2 | 73.1 | 26.4 | 59.2 |
| Ours (real only) | 84 | **81** | 83 | 80.2 | **43.2** | 92.4 | 1.0 | 46.0 | 35.4 | 74.8 | **27.9** | 59 |
| Ours (real+synth) | **85.7** | 80.3 | **85.2** | **83.2** | 40.5 | 92.7 | **29.7** | 42.8 | **38** | **75.9** | 22.6 | **61.6** |

There are many extensions that would be interesting. First, the presented ideas can be extended to produce continuous video streams in which each frame is densely annotated. Second, the presented approach can be applied to produce ground-truth data for many dense prediction problems, including optical flow, scene flow, depth estimation, boundary detection, stereo reconstruction, intrinsic image decomposition, visual odometry, and more. Third, our ideas can be extended to produce instance-level – rather than class-level – segmentations. There are many other exciting opportunities and we believe that modern game worlds can play a significant role in training artificial vision systems.

**Acknowledgements.** SRR was supported in part by the German Research Foundation (DFG) within the GRK 1362. Additionally, SRR and SR were supported in part by the European Research Council under the European Union's Seventh Framework Programme (FP/2007-2013) / ERC Grant Agreement No. 307942. Work on this project was conducted in part while SRR was an intern at Intel Labs.